\documentclass{article}
\usepackage{iclr2027_conference,times}
\iclrfinalcopy
\pdftrailerid{}

\usepackage{amsmath,amsfonts,bm}

\def\eqref#1{equation~\ref{#1}}

\def\1{\bm{1}}

\DeclareMathAlphabet{\mathsfit}{\encodingdefault}{\sfdefault}{m}{sl}
\SetMathAlphabet{\mathsfit}{bold}{\encodingdefault}{\sfdefault}{bx}{n}

\usepackage[utf8]{inputenc}
\usepackage[T1]{fontenc}
\usepackage{hyperref}
\usepackage{url}
\usepackage{booktabs}
\usepackage{amsmath,amssymb}
\usepackage{graphicx}
\usepackage{xcolor}
\usepackage{colortbl}
\usepackage{multirow}
\usepackage{tabularx}
\usepackage{wrapfig}
\usepackage{capt-of}
\usepackage{float}
\usepackage{enumitem}
\setlist[itemize]{leftmargin=1.5em,itemsep=1pt,topsep=2pt}
\usepackage{needspace}
\usepackage{microtype}

\definecolor{ourblue}{rgb}{0.368,0.507,0.71}
\hypersetup{colorlinks, linkcolor=ourblue, citecolor=ourblue, urlcolor=ourblue}
\definecolor{groupbg}{gray}{0.92}
\definecolor{groupyellow}{RGB}{255,246,210}
\definecolor{accentbg}{RGB}{229,238,250}
\newcolumntype{Y}{>{\centering\arraybackslash}X}

\title{\raggedright Where Activation Sparsity and\\ KV-Cache Sparsity Cross in LLM Decoding\vspace{8pt}}

\author{%
  \vspace*{4pt}
  \makebox[5.28in][c]{%
  \begin{tabular}{@{}c@{\hspace{0.02in}}c@{\hspace{0.02in}}c@{}}
    \parbox[c]{1.76in}{\centering Jungseob Lee$^{1}$\\[1pt] \fontsize{8pt}{7.5pt}\texttt{omanma1928@korea.ac.kr}} &\vspace*{4pt}
    \parbox[c]{1.76in}{\centering Seungyoon Lee$^{1}$\\[1pt] \fontsize{8pt}{7.5pt}\texttt{dltmddbs100@korea.ac.kr}} &\vspace*{4pt}
    \parbox[c]{1.76in}{\centering Seongtae Hong$^{1}$\\[1pt] \fontsize{8pt}{7.5pt}\texttt{ghdchlwls123@korea.ac.kr}}\\[4pt]\vspace*{4pt}
    \makebox[0pt][l]{\parbox[c]{1.76in}{\centering Sugyeong Eo$^{2,\dagger}$\\[1pt] \fontsize{8pt}{7.5pt}\texttt{s.eo@yonsei.ac.kr}}} &
    \makebox[0pt][l]{\parbox[c]{1.76in}{\centering Heuiseok Lim$^{1,\dagger}$\\[1pt] \fontsize{8pt}{7.5pt}\texttt{limhseok@korea.ac.kr}}} &
    \makebox[0pt][l]{}
  \end{tabular}%
  }\\[7pt]
  \small
  \makebox[5.28in][c]{%
    \begin{tabular}{c}
      $^{1}$Korea University \hspace{0.1in} $^{2}$Yonsei University Mirae Campus
    \end{tabular}%
  }%
}

\begin{document}

\maketitle
\begingroup
\renewcommand{\thefootnote}{\textdagger}
\footnotetext{Corresponding authors.}
\endgroup
\lhead{Preprint}
\vspace{-20pt}

\begin{abstract}
At each step, decoding one sequence with a large language model rereads the projection weights, whose traffic is fixed, and the key-value (KV) cache, whose traffic grows with context. Activation sparsity trims the first term and KV-cache sparsity the second, yet their reported speedups are hard to compare because each depends on context length and on the dense attention kernel it is measured against. We derive a byte crossover, the context length at which the two savings are equal, together with ideal speedup bounds for each branch and for their composition, from model dimensions and keep ratios alone. We then time both branches and their composition from 2K to 128K tokens on two GPUs after a dense prefill of real text, with dense and sparse modes reading the cache through the same split-K attention kernel. The projection branch leads at short context and the KV branch at long context, with speedups that follow their byte bounds up to fixed kernel costs. Adding these costs, measured in separate sweeps, lets the byte account predict the measured crossings of three keep-ratio pairs, a second model, and a second GPU to within 4.1K tokens. Timing the dense baseline with masked instead of split-K attention inflates the apparent speedup of the same KV policy about fivefold. An attention-scored KV selection answers the same passkey and multi-key placements as dense decoding up to 127K tokens, whereas a KV window misses most of them. Under matched perplexity budgets, activation sparsity composed with this selection decodes 14 to 26\% faster than the best single branch on both GPUs. Code is available at \url{https://github.com/js-lee-AI/ByteCross}.
\end{abstract}

\section{Introduction}
\label{sec:intro}

\begin{figure}[t]
\centering
\setlength{\abovecaptionskip}{6pt}
\includegraphics[width=\textwidth]{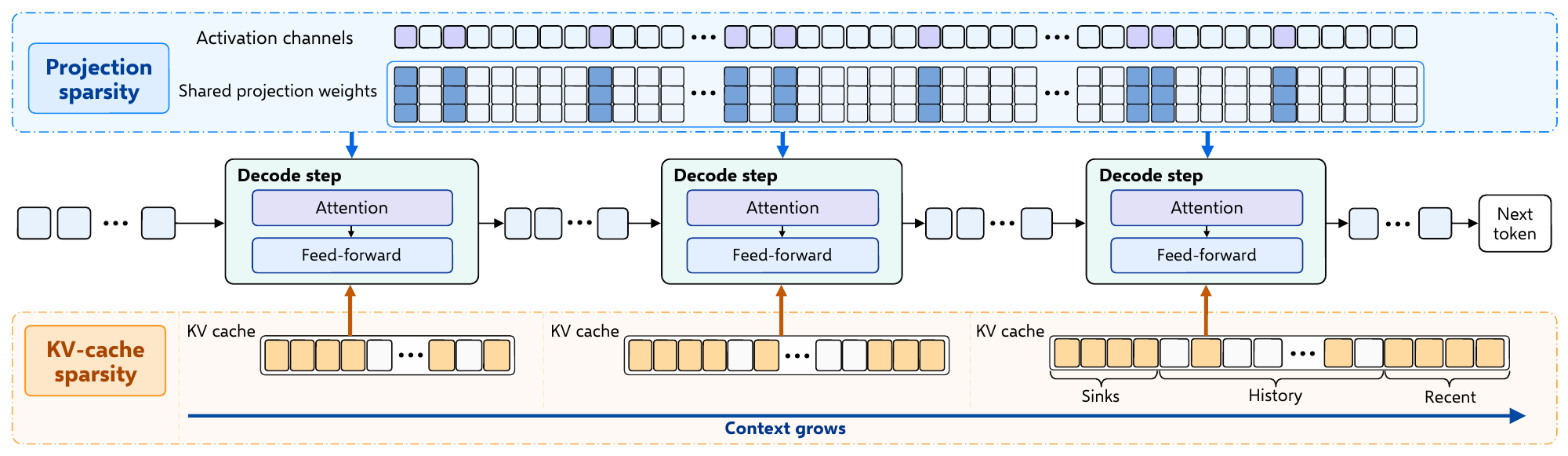}%
\caption{Decode-step reads of the two sparse branches as context grows. Blue marks the projection weights that the activations of each token select, and amber marks the KV entries that an attention-scored selection keeps, together with four sinks and the most recent entries. Unfilled cache tiles stay allocated but unread.}
\label{fig:hero}
\end{figure}

Decoding with a large language model (LLM) produces one token at a time, and for a single sequence each step is limited by memory bandwidth rather than arithmetic~\citep{fasttransformerdecoding2019,roofline,scalingtransformer2022,llmroofline2024}. Every step reads the projection weights of all layers and the key-value (KV) cache of all earlier tokens, so each byte it moves feeds roughly one floating-point operation. Serving systems raise hardware utilization through batching, paged cache storage, and fused attention kernels~\citep{vllm2023,orca2022,sglang2023,flashattention2022}, yet the traffic of a single decode step keeps the same two-part structure. As Figure~\ref{fig:hero} illustrates, projection traffic is fixed, whereas cache traffic grows linearly with context.

Two families of sparsity, which we refer to as the projection and KV branches, reduce these two terms. Activation sparsity skips the weights of channels whose activations are small~\citep{dejavu2023,teal2025,cats2024,powerinfer2023,prosparse2024,llmflash2023}, and KV-cache sparsity reads only part of the cache~\citep{h2o,streamingllm,snapkv,quest2024,specattn2026}. Because one saving is constant and the other grows with context, neither family removes more bytes at every length. However, the two lines of work report speedups at different context lengths, over different dense kernels, and at keep ratios with different quality costs, so their numbers cannot be compared directly. Choosing between the branches, or combining them, begins with the context length at which their savings become equal, but measured latency and a shared quality budget can both move that boundary.

We address this with a byte account of the decode step and a controlled timing harness. By counting the bytes each branch removes, we obtain a crossover length $n^*$, the context at which the two savings match, along with byte bounds on the speedups of both branches and their composition. Both depend only on model dimensions and keep ratios. We then time dense decoding, each branch alone, and both together on two GPUs at context lengths from 2K to 128K. Each of these decodes follows a dense prefill of real text, and all modes read their caches with one split-K attention kernel, which holds the dense baseline fixed. Our KV branch keeps the cache entries that an attention-scored selection picks once after prefill, gathered into one contiguous buffer, so each step reads only the retained bytes.

Short contexts favor the projection branch and long contexts the KV branch, as the byte account predicts, and the KV selection runs within 1\% of its byte bound on RTX A6000. Once the kernel cost of each branch is timed in separate sweeps, the account locates all six latency crossings we measured, and costs timed on A100 also locate the two RTX A6000 crossings. The dense attention kernel alone changes the apparent gain of the same KV policy about fivefold, because a slow baseline rewards any method that shortens the attended sequence. Likewise, a KV window can meet a perplexity budget and still fail to retrieve information outside its retained region, whereas the attention-scored selection matches dense decoding on every passkey and multi-key placement. Composed with this selection, activation sparsity is faster than either branch alone at matched perplexity. In summary, we make three contributions.
\begin{itemize}
\item A byte account of the decode step that gives, from model dimensions alone, the crossover of the two savings and the ideal speedup of each branch and of their composition, and that places every measured latency crossing on two models and two GPUs within 4.1K tokens once the kernel cost of each branch is added.
\item A timing protocol in which dense and sparse modes share one split-K attention kernel and decode after a real-text prefill, together with the finding that the attention path of the dense baseline moves the reported gain of the same KV policy at 128K from 1.29$\times$ to 6.92$\times$.
\item A quality-admissible composition of activation sparsity and an attention-scored KV selection, which decodes 25 to 26\% faster than the best single branch on RTX A6000 and 14 to 24\% faster on A100 at every perplexity budget that admits projection sparsity, and whose selection passes passkey and multi-key retrieval tests up to 127K tokens.
\end{itemize}

\section{Related Work}
\label{sec:related}

\paragraph{Memory-bound decoding and serving.}
Single-sequence decoding has an arithmetic intensity of about one, so its token latency tracks the bytes it moves~\citep{fasttransformerdecoding2019,roofline,scalingtransformer2022,llmroofline2024}. Serving systems attack the resulting under-utilization with paged KV storage and continuous batching~\citep{vllm2023,orca2022,sglang2023}, chunked prefills~\citep{sarathi2024}, fused attention and decoding kernels~\citep{flashattention2022,flashdecodingpp2023}, phase-disaggregated scheduling~\citep{splitwise2023,distserve2024}, and grouped-query attention, which shrinks the cache~\citep{gqa2023}. Weight-only quantization reduces the weight term uniformly~\citep{gptq2022,awq2023}. These techniques raise the throughput of dense decoding but keep its two-term structure, and the crossover decides which sparse mechanism, if any, runs on top of them.

\paragraph{Projection-side sparsity.}
Feed-forward activations are mostly near zero at inference time~\citep{lazyneuron2022}, and contextual sparsity predicts or thresholds the active channels and loads only their weights~\citep{dejavu2023,relustrikesback2023,turbosparse2024,prosparse2024,qsparse2024}. Training-free variants threshold activation magnitudes directly~\citep{cats2024,teal2025,rsparse2025,duogpt2025}, predictor-based variants learn which neurons fire~\citep{shadowllm2024}, offloading engines exploit the same locality across the memory hierarchy~\citep{powerinfer2023,llmflash2023}, and batched variants restrict sparsity to the components that stay sparse under batching~\citep{polarsparsity2025}. All of them reduce the fixed projection term, and we ask how their benefit compares with cache-read sparsity as context grows.

\paragraph{KV-cache sparsity.}
Methods for the growing term evict entries by attention scores~\citep{h2o,scissorhands2023,fastgen2023,pyramidkv2024,tova2024}, keep attention sinks plus a recent window~\citep{streamingllm,duoattention2024}, compress the prompt cache before generation~\citep{snapkv}, select query-relevant blocks in prefill or decode~\citep{quest2024,infllm2024,retrievalattention2024,minference2024}, train sparse attention natively~\citep{nsa2025,moba2025}, verify sparse-attention drafts speculatively~\citep{specattn2026,sparsespec2025}, or quantize the cache~\citep{kivi2024}. In our byte account, each of these methods enters through its effective keep ratio, while selection, packing, and cache updates add costs of their own. Our KV branch applies a SnapKV-style selection once after prefill and gathers the retained entries into a contiguous buffer, so its decode step pays only for the entries it reads.

\paragraph{Execution-path methods.}
Speculative decoding and its self-speculative and cascade variants reduce the number of full-model steps~\citep{specdecoding2022,medusa2024,eagle2024,draftverify2023,layerskip2024,fastercascades2025,lee2026dart_routing,casspec2025}, adaptive layer selection chooses which blocks to run~\citep{knapspec2026}, and both compose with either branch, since each branch reduces only the bytes a step reads. To our knowledge, no prior work measures projection-side and KV-cache sparsity against each other along the context axis.

\section{The Byte Crossover}
\label{sec:method}

We count the reads each branch removes from a decode step, equate the two savings to obtain the crossover length, and derive ideal speedup bounds. Throughout the paper, we write K for 1,024 tokens and use decimal GB and MB for bandwidth and storage.

\subsection{Decode-Step Memory Traffic}
\label{sec:traffic}

At batch size 1, a decode step performs about one floating-point operation (FLOP) per byte it reads, two orders of magnitude below the compute-bound regime of an A100~\citep{roofline,fasttransformerdecoding2019}. Reads from high-bandwidth memory (HBM) therefore set its latency, so we compare the two branches by the bytes they remove. At context length $n$, we write the read traffic of one step as
\begin{equation}
B_\text{total}(n) =
\underbrace{B_\text{MLP}}_{\text{constant}} +
\underbrace{B_\text{Attn}}_{\text{constant}} +
\underbrace{B_\text{KV}(n)}_{\text{linear in } n} +
\underbrace{B_\text{other}}_{\text{constant}},
\label{eq:bandwidth}
\end{equation}
where $B_\text{MLP}$ and $B_\text{Attn}$ are the multilayer perceptron (MLP) and attention projection weights that each decode step loads once, $B_\text{KV}(n)$ is the attention read over stored keys and values, and $B_\text{other}$ covers the language-model head, one embedding row, and normalization weights. Only $B_\text{KV}(n)$ grows with context.

Because the layer count cancels when we equate the two savings, one-layer counts suffice. Let $d$ be the hidden size, $d_\text{ff}$ the feed-forward size, $h_\text{kv}$ the number of KV heads, and $d_h$ the head dimension. In 16-bit precision, one layer reads
\begin{align}
B_\text{MLP} &= 3 \cdot d \cdot d_\text{ff} \cdot 2 \text{ bytes}, \label{eq:bmlp} \\
B_\text{Attn} &= (2d^2 + 2 \cdot d \cdot h_\text{kv} \cdot d_h) \cdot 2 \text{ bytes}, \label{eq:battn} \\
B_\text{KV}(n) &= 2 \cdot h_\text{kv} \cdot d_h \cdot n \cdot 2 \text{ bytes}. \label{eq:bkv}
\end{align}
For scale, one Llama-3.1-8B layer reads 436~MB of projection weights at every step, while its cache adds $4096n$ bytes, 2~MB at $n{=}512$ and 134~MB at $n{=}32$K. Across a whole Qwen3-8B step, profiled in Appendix~\ref{app:traffic}, the cache grows from 0.5\% of all reads at 512 tokens to 24.2\% at 32K, while the feed-forward weights alone still account for more than half.

\needspace{8\baselineskip}
\subsection{Sparse Branches}
\label{sec:branches}

\begin{wrapfigure}{r}{0.55\textwidth}
\vspace{-0.45cm}
\centering
\includegraphics[width=0.54\textwidth]{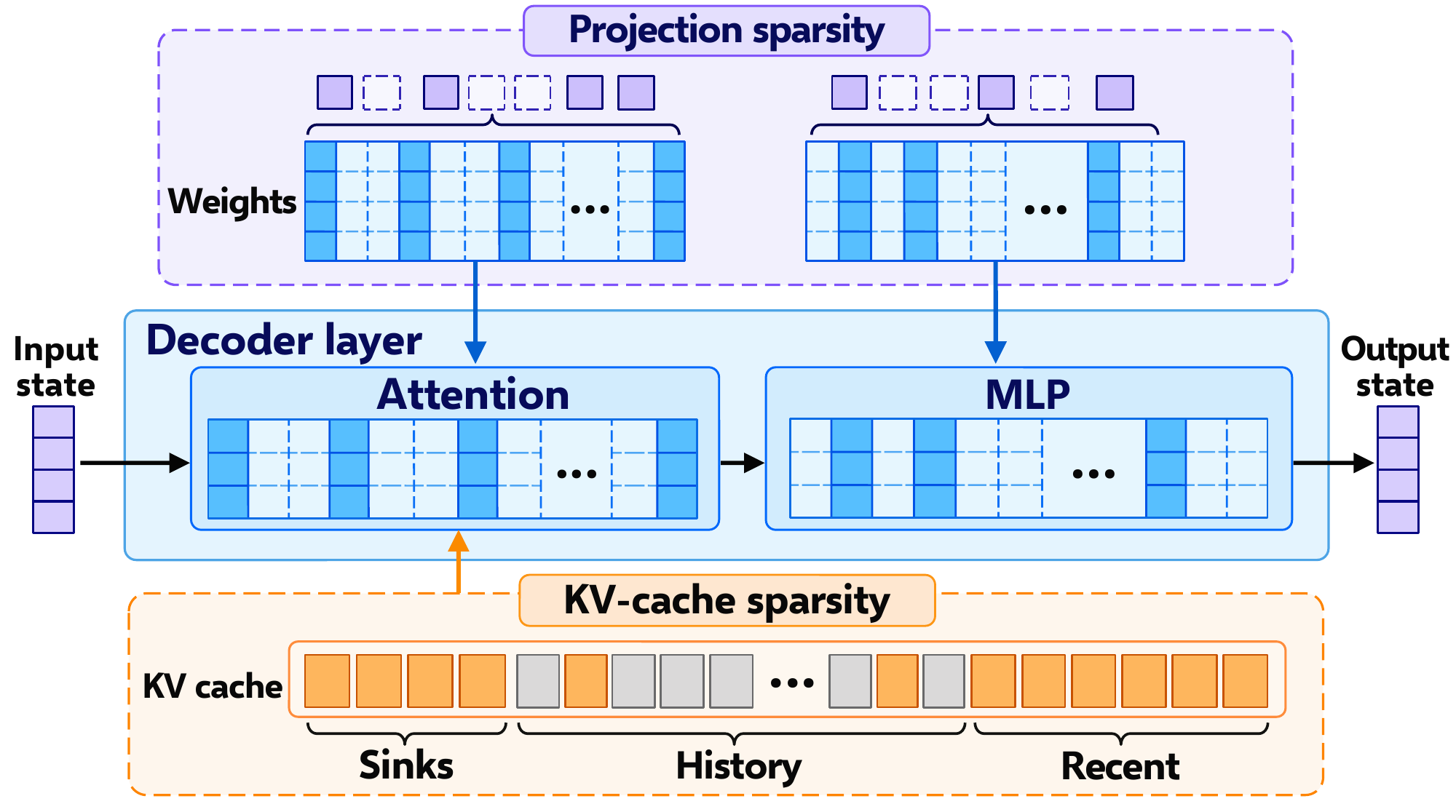}
\vspace{-0.2cm}
\caption{Read paths inside one decoder layer. Blue marks activation-selected projection reads and amber the retained KV reads, which serve attention only. Gray cache entries stay allocated but unread.}
\label{fig:read_paths}
\vspace{-0.3cm}
\end{wrapfigure}
Two in-layer mechanisms remove the two removable terms, and Figure~\ref{fig:read_paths} shows where each acts inside a decoder layer. Because both leave every layer in place, we can compare their savings before measuring any latency.

\textbf{Projection-side sparsity} keeps a fraction $r_\text{P}$ of the activation channels and loads only the associated weights, as threshold-based activation sparsity does~\citep{teal2025,cats2024}. We sparsify all seven projections of a layer, the three feed-forward and the four attention projections, so one layer saves $\Delta B_\text{P} = (B_\text{MLP} + B_\text{Attn})(1 - r_\text{P})$ bytes independently of context length.

\textbf{KV-cache sparsity} reads a fraction $r_\text{KV}$ of the stored keys and values, whether the retained entries come from a sink-plus-recent window~\citep{streamingllm}, accumulated attention scores~\citep{h2o}, or query-aware retrieval~\citep{quest2024,specattn2026}. One layer then saves $\Delta B_\text{KV}(n) = B_\text{KV}(n)(1 - r_\text{KV})$ bytes, which grows linearly with $n$.

In the form we study, neither branch discards state, and each changes only what a step reads. Unselected weights stay in memory, and the full cache stays allocated and keeps receiving new entries. A system can switch between the branches from one step to the next without rebuilding any state.

\subsection{Crossover Length and Speedup Bounds}
\label{sec:rule}

\begin{wraptable}{r}{0.56\textwidth}
\vspace{-0.4cm}
\centering
\caption{Byte crossover lengths at 50\% projection keep and 30\% KV keep, for all seven projections and for the feed-forward projections alone. Window is the released context length.}
\label{tab:crossover}
\small
\setlength{\tabcolsep}{3pt}
\begin{tabular}{lccccc}
\toprule
\rowcolor{groupbg}
\textbf{Model} & $d$ & $d_\text{ff}$ & $n^*$ & $n^*_\text{FF}$ & \textbf{Window} \\
\midrule
Qwen3-8B & 4096 & 12288 & 65.7K & 51.4K & 40K \\
Llama-3.1-8B & 4096 & 14336 & 74.3K & 60.0K & 128K \\
Mistral-7B & 4096 & 14336 & 74.3K & 60.0K & 32K \\
Llama-3.1-70B & 8192 & 28672 & 291.4K & 240.0K & 128K \\
\bottomrule
\end{tabular}
\vspace{-0.3cm}
\end{wraptable}
We define the \textit{crossover length} $n^*$ as the context at which the two savings match, $\Delta B_\text{P} = \Delta B_\text{KV}(n^*)$. With $P = 3 d d_\text{ff} + 2d^2 + 2 d h_\text{kv} d_h$ parameters in the seven projections of a layer, substituting Equations~\ref{eq:bmlp} to~\ref{eq:bkv} and cancelling the shared element size gives
\begin{equation}
n^* = \frac{P \cdot (1 - r_\text{P})}{2 \cdot h_\text{kv} \cdot d_h \cdot (1 - r_\text{KV})}.
\label{eq:crossover}
\end{equation}
Equation~\ref{eq:crossover} is a byte threshold with no fitted parameter. Below $n^*$ the projection branch removes more bytes, and above it the KV branch does. For a branch that sparsifies only the feed-forward projections, $P$ reduces to $3 d d_\text{ff}$ and the crossover moves earlier. Table~\ref{tab:crossover} lists both thresholds for four models, all with eight KV heads of dimension $d_h{=}128$. Among the 8B models the one with the smaller feed-forward layer crosses earlier, while the 70B model crosses far beyond its 128K window, so at these keep ratios projection sparsity removes more of its bytes at every supported length. The KV layout moves the crossover about as much as model size does, and with 32 KV heads Llama-3.1-8B would cross near 20.7K.

On a step that is purely bandwidth-bound, a branch that removes $\Delta B$ of the $B_\text{total}(n)$ bytes can speed decoding up by at most $B_\text{total}(n)/(B_\text{total}(n)-\Delta B)$. This bound falls with $n$ for the projection branch, rises for the KV branch, and is largest for the two together. The crossover and the three bounds are the predictions we test. The threshold also scales with precision. With weights of $s_\text{w}$ bytes and cache elements of $s_\text{kv}$ bytes, it is multiplied by $s_\text{w}/s_\text{kv}$, so 4-bit weights with a 16-bit cache move the crossover four times closer.

\textbf{Dispatch and batching.} The byte crossover indicates which branch removes more traffic, but fixed kernel costs can move the latency crossing. We therefore also define a \textit{latency gate}, which times dense decoding and both branches at representative lengths and selects for each request the fastest admissible configuration from the resulting lookup table, one-time costs included, where admissibility also requires quality tests. For a KV method with a fixed token budget $b$, we substitute $r_\text{KV}=b/n$ and obtain $n^* = b + P(1-r_\text{P})/(2h_\text{kv}d_h)$ whenever the crossing lies above $b$. Finally, a batch of $B$ sequences shares each weight read while every sequence reads its own cache. A sparse projection then reads every channel that any sequence of the batch activates, a fraction $u_B \geq r_\text{P}$ of its weights, so the crossover of each sequence becomes $n^*_B = P(1-u_B)/\bigl(2Bh_\text{kv}d_h(1-r_\text{KV})\bigr)$, at most $n^*/B$.

\section{Where the Branches Cross}
\label{sec:results}

\subsection{Setup}
\label{sec:setup}

\textbf{Models and branches.} We use Llama-3.1-8B-Instruct~\citep{llama3} as the primary model because its 128K window spans its predicted crossover. Llama-3.2-3B-Instruct~\citep{llama3} tests the crossover on a smaller layer, Qwen3-8B~\citep{qwen3} provides a traffic profile inside its 40K window, and with Mistral-7B-v0.3~\citep{mistral7b} we check transfer to another checkpoint. For the projection branch we use the training-free activation sparsity of TEAL~\citep{teal2025}. It selects activation channels with magnitude thresholds calibrated on Llama-3-8B, loads only the corresponding weights through a Triton~\citep{triton2019} sparse matrix-vector kernel, and targets 50\% keep in all seven projections, reading 51\% of the projection bytes on real-text decode steps. Thresholds recalibrated on the evaluated model lower every perplexity increase slightly, so the Llama-3-8B thresholds are a conservative choice, as Appendix~\ref{app:details} shows.

Our KV branch follows SnapKV~\citep{snapkv}. After dense prefill it scores the cache once with the attention of the last 64 prompt tokens, keeps four sink tokens and the highest-scoring entries of each KV head up to a 30\% keep ratio, and gathers them into a contiguous buffer that every later step reads in one kernel call and extends with its own key and value. On A100 the one-time scoring and gather take 0.24~s for a 32K-token prompt, which the faster decode steps repay within 146 generated tokens at every length from 16K to 128K. For comparison we also time a sink-plus-recent window in the style of attention sinks~\citep{streamingllm}, which reads as many entries in place.

\textbf{Timing and quality.} We time all modes in FP16 at batch size 1 with compiled decoding, and dense and sparse modes alike read their caches through the split-K decoding kernel of the FlashAttention library~\citep{flashattention2022}. For each context length we first prefill the cache densely with that many tokens of WikiText-2 text and then time repeated 50-step decodes, so every sparse branch sees the activations of real text. Each cell of our main campaigns runs in a fresh process with full-graph compilation, and every campaign repeats in three to five blocks on one GPU type. Timing uses A100-SXM4 80~GB and RTX A6000 48~GB GPUs, and on RTX A6000 the sweep ends at 64K because its 128K cells do not fit in memory.

We measure quality on real text with decode-position perplexity (PPL) on WikiText-2~\citep{wikitext2}, which prefills densely and scores 512 tokens one at a time under the branch, and with passkey and multi-key retrieval tests from 32K to 127K tokens. Appendix~\ref{app:details} gives the full protocols, and Appendix~\ref{app:projection} reports the speed and task-level quality of the projection branch at short context.

\subsection{Speedup Across Context}
\label{sec:sweep}

\begin{figure}[t]
\centering
\setlength{\abovecaptionskip}{6pt}
\includegraphics[width=\textwidth]{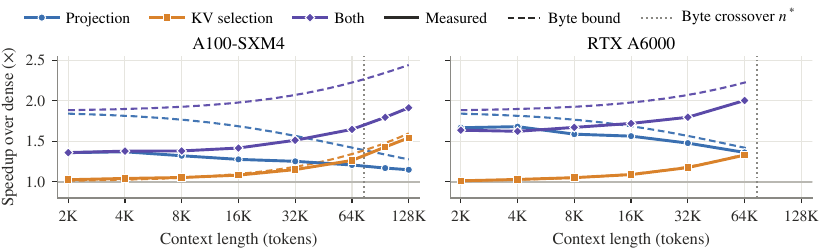}%
\caption{Speedup over dense decoding of the projection branch, the KV selection, and both together on Llama-3.1-8B, after a dense prefill of real text. Solid lines are measured, dashed lines are byte bounds, and the dotted line marks the byte crossover. RTX A6000 runs stop at 64K.}
\label{fig:sweep}
\end{figure}

Figure~\ref{fig:sweep} shows the speedup of both branches and their composition over dense decoding from 2K to 128K, beside the byte bound of each. The projection branch leads at short context, reaching 1.36$\times$ on A100 and 1.67$\times$ on RTX A6000 at 2K, and its advantage shrinks as cache reads take a growing share of the step. The KV selection follows the opposite trend, overtakes the projection branch between 32K and 64K on A100, and reaches 1.54$\times$ at 128K. Because both trends mirror their bounds, the byte account alone predicts which branch leads at short and long context.

The selection runs within 1\% of its bound on RTX A6000 and within 6\% on A100, and the projection branch comes within 4 to 10\% of its bound on RTX A6000. The remaining gap is a kernel cost that stays nearly constant across context, about 2~ms of step time for the sparse projection on A100 and 1 to 1.5~ms on RTX A6000, which Section~\ref{sec:crossing} uses to predict the crossings.

Composition has the highest bound at every length, and from 8K onward it is the fastest mode on both GPUs. It reaches 1.91$\times$ at 128K on A100 and 2.00$\times$ at 64K on RTX A6000, against 1.54$\times$ and 1.36$\times$ for the faster single branch.

\subsection{Locating the Crossing}
\label{sec:crossing}

\begin{figure}[t]
\centering
\setlength{\abovecaptionskip}{6pt}
\includegraphics[width=\textwidth]{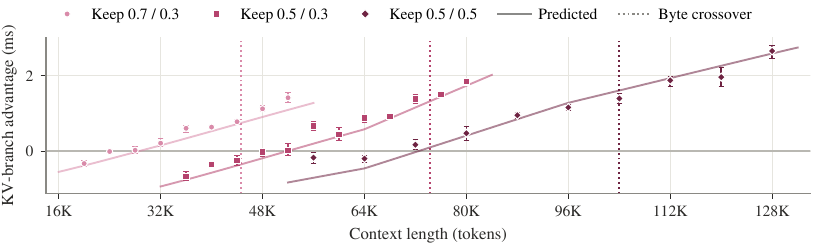}%
\caption{KV-selection advantage, the projection minus the selection step latency, on A100-SXM4. Points average five fresh-process blocks with 95\% intervals, lines are predicted without any crossing cell, and dotted lines mark byte crossovers.}
\label{fig:crossing}
\end{figure}

\begin{wraptable}{r}{0.47\textwidth}
\vspace{-0.4cm}
\centering
\caption{Byte crossover $n^*$, cost-aware prediction, and measured crossing, in K tokens. Each crossing is measured in fresh-process blocks, five on A100-SXM4 and four on RTX A6000.}
\label{tab:crossings}
\small
\setlength{\tabcolsep}{5pt}
\begin{tabular}{lcccc}
\toprule
\rowcolor{groupbg}
\textbf{Model} & $r_\text{P}\,/\,r_\text{KV}$ & $n^*$ & \textbf{Pred.} & \textbf{Measured} \\
\midrule
\rowcolor{groupyellow}\multicolumn{5}{c}{\textit{A100-SXM4}\strut} \\
8B & 0.7\,/\,0.3 & 44.6 & 28.6 & 25.1 \\
8B & 0.5\,/\,0.3 & 74.3 & 51.9 & 50.4 \\
8B & 0.5\,/\,0.5 & 104.0 & 72.4 & 68.3 \\
3B & 0.5\,/\,0.3 & 34.3 & 13.9 & 16.5 \\
\rowcolor{groupyellow}\multicolumn{5}{c}{\textit{RTX A6000}\strut} \\
8B & 0.7\,/\,0.3 & 44.6 & 38.7 & 39.3 \\
8B & 0.5\,/\,0.3 & 74.3 & 68.5 & 68.7 \\
\bottomrule
\end{tabular}
\vspace{-0.3cm}
\end{wraptable}
The byte crossover assumes that both branches run at their byte bounds. When each branch also pays a kernel cost beyond its byte time, $c_\text{P}$ or $c_\text{KV}$, the latencies cross near $n_\text{lat} \approx n^* - (c_\text{P} - c_\text{KV})/s$, where $s$ is the growth of the byte-time gap as the context gains 1K tokens, the KV bytes that the KV branch skips within those tokens divided by the effective bandwidth. On A100, $s$ is only about 0.06~ms, so a kernel-cost difference of one millisecond moves the crossing by roughly 17K tokens. We therefore predict each crossing from the byte account together with the kernel cost of each branch, following the procedure of Appendix~\ref{app:harness}. These costs come from the context sweep of Figure~\ref{fig:sweep}, a 32K keep-ratio campaign, and a sweep of Llama-3.2-3B over five context lengths, so that no cell of the crossing campaigns enters any prediction.

Figure~\ref{fig:crossing} and Table~\ref{tab:crossings} compare these predictions with the measured crossings. On A100 every prediction falls within 4.1K tokens of its crossing, for the three keep-ratio pairs of Llama-3.1-8B as well as for the smaller layer of Llama-3.2-3B. The same account carries over to a GPU with less bandwidth. On RTX A6000, $s$ is about 0.13~ms, so a millisecond of kernel cost moves the crossing only about 8K tokens, and both crossings fall within 0.6K tokens of their predictions. Kernel costs measured on A100, combined with the dense step time of the RTX A6000, also place these two crossings within 3.5K tokens of the measurements.

Across both models and both GPUs the crossings keep the order of their byte crossovers, and how far each falls below $n^*$ depends on the kernel cost of each branch relative to the memory bandwidth of the GPU. Appendix~\ref{app:replication} lists every context of the campaigns and how each crossing is located.

\subsection{The Attention Baseline}
\label{sec:paths}
\suppressfloats[t]

\begin{wrapfigure}{r}{0.5\textwidth}
\vspace{-0.5cm}
\centering
\includegraphics[width=0.49\textwidth]{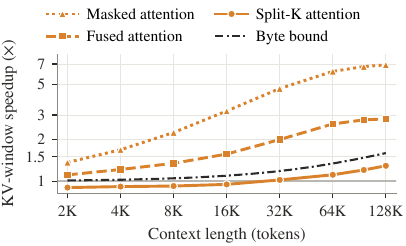}
\vspace{-0.25cm}
\caption{Speedup of the same 30\% KV window over dense decoding on A100 when both read the cache through masked, fused, or split-K attention. Only split-K attention keeps the speedup below the byte bound of the window.}
\label{fig:paths}
\vspace{-0.35cm}
\end{wrapfigure}
Figure~\ref{fig:paths} times the same KV window against dense decoding with masked, fused, or split-K attention, and Appendix~\ref{app:harness} lists the throughput of each path. At 128K the window runs 6.92$\times$ faster than dense decoding with masked attention and 2.82$\times$ faster with fused attention, whereas split-K attention leaves it 1.29$\times$ faster, below its byte bound of 1.60$\times$. The first two gains exceed the bound because the dense baselines themselves are inefficient. At 128K, masked attention uses 5\% of peak bandwidth and fused attention 12\%, against 78\% for split-K attention, so shortening the attended sequence removes wasted execution as well as cache bytes. A reported KV speedup therefore needs the attention path and execution mode of its baseline, and a gain above the byte bound indicates wasted execution in the dense baseline.

\textbf{Transfer.} Our projection branch carries over to another checkpoint. At 50\% keep it raises the decode throughput of Mistral-7B-v0.3 on RTX A6000 from 48.0 to 82.2~tok/s, a 1.71$\times$ speedup, as large as that of Llama-3.1-8B on the same GPU.

\textbf{Batching.} Larger batches move the crossover toward short context, as Appendix~\ref{app:batch} shows. On WikiText-2 decode steps of Llama-3.1-8B, the channels that at least one sequence of a batch activates cover 92\% of the projection weights at four sequences and 98.5\% at eight, so the batched crossover $n^*_4$ falls to 3.0K tokens, far below $n^*/4$.

\section{Quality, Composition, and Dispatch}
\label{sec:quality}

\subsection{Retrieval as an Admissibility Test}
\label{sec:kvquality}

\begin{wraptable}{r}{0.55\textwidth}
\vspace{-0.4cm}
\centering
\caption{Perplexity increase $\Delta$PPL over dense decoding at 32K and retrieval accuracy on Llama-3.1-8B-Instruct. Dense perplexity is 6.09.}
\label{tab:kvquality}
\scriptsize
\setlength{\tabcolsep}{2.5pt}
\begin{tabular}{lccccccc}
\toprule
\rowcolor{groupbg}
& & \multicolumn{3}{c}{\textbf{Passkey}} & \multicolumn{3}{c}{\textbf{Multi-key}} \\
\rowcolor{groupbg}
\multirow{-2}{*}{\textbf{Configuration}} & \multirow{-2}{*}{$\Delta$\textbf{PPL}} & 32K & 64K & 127K & 32K & 64K & 127K \\
\midrule
Dense & 0 & 40/40 & 40/40 & 40/40 & 40/40 & 40/40 & 36/40 \\
KV window, 30\% & 0.06 & 16/40 & 16/40 & 16/40 & 16/40 & 16/40 & 14/40 \\
Selection, 30\% & $<$0.01 & 40/40 & 40/40 & 40/40 & 40/40 & 40/40 & 36/40 \\
Selection, 20\% & $<$0.01 & 40/40 & 40/40 & 40/40 & 40/40 & 40/40 & 36/40 \\
Proj.\ 50\% + sel.\ 20\% & 0.96 & 40/40 & 40/40 & 40/40 & 40/40 & 40/40 & 36/40 \\
\bottomrule
\end{tabular}
\vspace{-0.3cm}
\end{wraptable}
Table~\ref{tab:kvquality} compares dense decoding with the two KV branches and with the selection composed with the projection branch. By perplexity at 32K, both KV branches look nearly free, whereas 50\% projection keep adds about one point. The selection also stays within 0.01 of dense perplexity at 64K and 127K.

In the passkey test, however, the window answers only the 16 placements inside its retained region at every length. In the multi-key test, which hides four similar access codes in natural text and asks for one of them, it again answers only placements inside that region and returns a distractor code nine times at 32K and 19 times at 127K. The selection answers every placement of both tests at 32K and 64K, alone and composed with the projection branch, and at 127K, where dense decoding itself misses four multi-key placements, it answers exactly the placements that dense decoding answers. When a second turn asks for the code of another archive, the selection scored for the first question answers all 40 placements at 32K and 64K, as does a selection rescored for the new question.

The 30\% window keeps more entries than the 20\% selection and still misses most placements. We therefore admit a KV configuration only if it also passes both retrieval tests. Appendix~\ref{app:longquality} details the long-context and two-turn evaluations.

\subsection{The Quality-Matched Crossover}
\label{sec:iso}

\begin{figure}[t]
\centering
\setlength{\abovecaptionskip}{6pt}
\includegraphics[width=\textwidth]{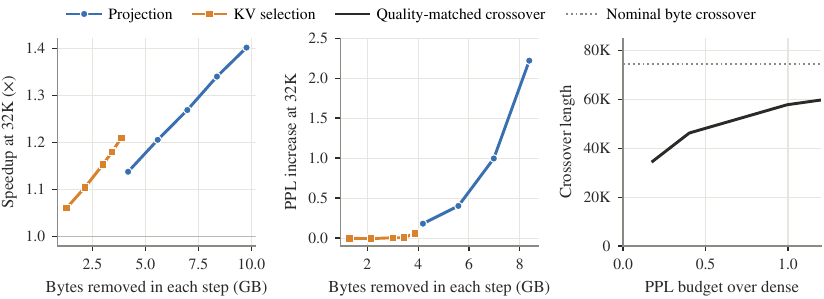}%
\caption{Keep-ratio sweeps at 32K. Left and middle, A100 speedup and decode-position PPL increase of the projection branch and the KV selection against the bytes each removes from a decode step. Right, the byte crossover at equal PPL increase for each budget.}
\label{fig:iso}
\end{figure}

Figure~\ref{fig:iso} sweeps the keep ratio of each branch at 32K, because our nominal keep ratios compare branches of unequal quality. Over the eight windows, the 95\% intervals of the perplexity increase separate the projection keep ratios of 70\%, 60\%, and 50\%, whereas the KV selection changes perplexity by less than 0.01 down to 20\% keep and by 0.06 at 10\%, where it still passes both retrieval tests. On A100 the selection reaches 1.21$\times$ at 10\% keep, within 4\% of its byte bound, against 1.27$\times$ for the projection branch at 50\% keep, which costs about one point of perplexity. Equating the byte savings of the two branches at equal PPL increase moves the crossover to 35.7K, 48.1K, and 57.8K for budgets of 0.2, 0.5, and 1.0, well below the nominal 74.3K.

\subsection{Composition Under a Quality Budget}
\label{sec:composition}

\begin{wraptable}{r}{0.5\textwidth}
\vspace{-0.4cm}
\centering
\caption{Fastest configuration within each perplexity budget at 32K on RTX A6000, timed in five independent blocks. The KV branch is the attention-scored selection, a keep ratio of 1 disables a branch, and shaded rows select a composition.}
\label{tab:composition}
\small
\setlength{\tabcolsep}{4pt}
\begin{tabular}{ccccc}
\toprule
\rowcolor{groupbg}
& & & \multicolumn{2}{c}{\textbf{Speedup over}} \\
\rowcolor{groupbg}
\multirow{-2}{*}{\textbf{Budget}} & \multirow{-2}{*}{$r_\text{P}\,/\,r_\text{KV}$} & \multirow{-2}{*}{$\Delta$\textbf{PPL}} & \textbf{dense} & \textbf{best single} \\
\midrule
0.1 & 1.0\,/\,0.2 & 0.004 & 1.20$\times$ & 1.00$\times$ \\
\rowcolor{accentbg}
0.2 & 0.7\,/\,0.2 & 0.146 & 1.53$\times$ & 1.25$\times$ \\
\rowcolor{accentbg}
0.5 & 0.6\,/\,0.2 & 0.402 & 1.69$\times$ & 1.26$\times$ \\
\rowcolor{accentbg}
1.0 & 0.5\,/\,0.2 & 0.963 & 1.86$\times$ & 1.25$\times$ \\
\bottomrule
\end{tabular}
\vspace{-0.3cm}
\end{wraptable}
Table~\ref{tab:composition} tests whether composition keeps its advantage once each configuration must meet a perplexity budget. We evaluate 16 dense, single-branch, and composed settings on eight fixed 32K WikiText-2 windows and report all of them with bootstrap intervals in Appendix~\ref{app:composition}. Within each budget, a timing screen selects the fastest configuration and the fastest single branch, and the selected settings are re-timed in five independent blocks.

From a budget of 0.2 upward, the selected compositions run 25 to 26\% faster than the best single branch, and each of the five blocks shows a gain of at least 21\%. The same compositions also lead on A100, by 14 to 24\% over the best single branch. At the tightest budget of 0.1, which no projection keep ratio of the grid meets, the selection alone is the fastest configuration and still decodes 1.20$\times$ faster than dense decoding. Because the selection changes perplexity by less than 0.01 at every keep ratio of the grid, each composition pairs 20\% KV keep with the most projection sparsity that its budget admits. Every selected configuration also passes both retrieval tests.

\subsection{Dispatch on Request Mixes}
\label{sec:trace}

\begin{wraptable}{r}{0.5\textwidth}
\vspace{-0.4cm}
\centering
\caption{Decode speedup over dense decoding of each dispatch policy on the request mix. Requests that use the selection also pay for its scoring pass.}
\label{tab:dispatch}
\small
\setlength{\tabcolsep}{4pt}
\begin{tabular}{lcc}
\toprule
\rowcolor{groupbg}
\textbf{Policy} & \textbf{A100-SXM4} & \textbf{RTX A6000} \\
\midrule
Always projection & 1.270 & 1.518 \\
Always KV selection & 1.088 & 1.101 \\
Always both & 1.401 & 1.694 \\
Calibrated gate & 1.404 & 1.716 \\
\bottomrule
\end{tabular}
\end{wraptable}
We also price a synthetic request mix that combines chat, long-document question answering~\citep{lee2026madara}, and agent loops~\citep{lee2026agentloop}, integrating the measured decode cost of each request along its generated sequence and charging the selection pass as a one-time cost of its prompt. On this mix, a gate calibrated on measured latency reaches 1.40$\times$ over dense decoding on A100 and 1.72$\times$ on RTX A6000, against 1.27$\times$ and 1.52$\times$ for always using the projection branch. Table~\ref{tab:dispatch} compares the dispatch policies, and Appendix~\ref{app:dispatch} specifies the mix. Always composing the two branches already reaches 1.40$\times$ and 1.69$\times$, so nearly all of the gate's gain comes from composition.

\section{Implications for Deployment}
\label{sec:implications}

For a deployment, the measurements suggest a short procedure. First, compute $n^*$ from the model dimensions at keep ratios that meet the target quality, since the quality-matched crossover can lie tens of thousands of tokens below the nominal one. Next, time dense decoding, both branches, and their composition at a few lengths on the deployed attention kernel, and add the kernel cost of each branch to its byte time to locate the latency crossing, because on A100 a kernel-cost difference of one millisecond moves that crossing by roughly 17K tokens. Finally, admit a KV configuration only after retrieval tests, because perplexity budgets do not detect the loss of distant information.

The account also shows how the balance shifts beyond batch size one. A batch shares each weight read, and the channels its sequences activate together leave a sparse projection little to skip, so in batched serving the crossover of each sequence falls to a few thousand tokens and the byte account favors KV-cache sparsity from short contexts onward. Weight quantization moves the crossover closer in the same way, whereas larger feed-forward layers push it outward, beyond the 128K window of Llama-3.1-70B at the keep ratios we study.

\section{Conclusion}
\label{sec:conclusion}

Projection-side and KV-cache sparsity remove different parts of decode traffic, and a byte account of the two parts predicts where their savings cross and how far the branches, alone or composed, can speed up decoding. On a common attention kernel and after a real-text prefill, the measured speedups track their bounds up to a fixed kernel cost of each branch, and with these costs included, the account predicts the latency crossing of every keep-ratio pair, model, and GPU we measured to within 4.1K tokens. The same account separates real savings from artifacts of a slow dense baseline and shows that a quality budget moves the crossover earlier. An attention-scored KV selection passes the passkey and multi-key tests up to 127K tokens that a KV window fails, and composed with activation sparsity it is the fastest configuration we measured under every perplexity budget that admits projection sparsity. We recommend reporting each sparse decoding speedup with its byte bound and the attention path of its dense baseline, and testing retrieval before admitting a KV configuration.

\subsection*{Ethics Statement}

This work measures the inference efficiency of publicly released language models on public benchmarks and synthetic prompts. It involves no human subjects, personal data, or new datasets. Faster decoding lowers the energy and hardware cost of serving existing models without changing what those models generate, so we foresee no risks beyond those of the models themselves.

\setlength{\bibsep}{7pt plus 3pt minus 2pt}
\bibliography{references}
\bibliographystyle{iclr2027_conference}

\newpage
\appendix
\raggedbottom

\section{Experimental Details}
\label{app:details}

\textbf{Model dimensions.} Llama-3.1-8B-Instruct has 32 layers, $d{=}4096$, $d_\text{ff}{=}14336$, eight KV heads with $d_h{=}128$, and a 128{,}256-token vocabulary. Llama-3.2-3B-Instruct has 28 layers, $d{=}3072$, $d_\text{ff}{=}8192$, eight KV heads with $d_h{=}128$, and the same vocabulary, so its byte crossover at 50\% projection keep and 30\% KV keep is 35{,}109 tokens, or 34.3K. Qwen3-8B has 36 layers, $d{=}4096$, $d_\text{ff}{=}12288$, eight KV heads with $d_h{=}128$, and a 151{,}936-token vocabulary. Mistral-7B-v0.3 has 32 layers, $d{=}4096$, $d_\text{ff}{=}14336$, eight KV heads, and $d_h{=}128$, so it shares the KV layout of Llama-3.1-8B.

\textbf{Software.} The projection branch applies the public TEAL implementation~\citep{teal2025} with Llama-3-8B magnitude thresholds inside the \texttt{gpt-fast} decoding loop with \texttt{torch.compile}, PyTorch~2.4, and CUDA~12.1. The FF-only variant patches only the gate, up, and down projections. The Mistral check reuses the projection-branch stack unchanged.

Decode attention in every mode is \texttt{flash\_attn\_with\_kvcache} from flash-attn~2.6.3, which reads the cache with split-K parallelism and is wrapped as an opaque custom operator so that compilation proceeds without a graph break at the attention call. The selection gathers its retained entries of each KV head once into a contiguous buffer, and every decode step appends its own key and value to that buffer and reads it in one call of the same kernel, while the full cache stays allocated and still receives each new entry. The window reads the full cache in place. Its kernel call starts at the first retained recent entry through a left-padding offset, and the four sink tokens form a second call over the cache head merged by log-sum-exp, so no cache entry is copied.

\textbf{Timing.} Every cell of the sweep, crossing, keep-ratio, and composition campaigns starts a fresh process, allocates the cache for its context length and the generated tokens, fills it by a dense prefill of the first $n$ tokens of the WikiText-2 test stream in chunks of 512 with causal FlashAttention, and then decodes from token $n$ of the same stream. It requires full-graph compilation and times five repeats of 50 steps after five warm-up steps. Each block holds every cell of its campaign on one GPU, in a varied order, and speedups are paired ratios to the dense cell of the same block. The sweeps use three blocks, the composition campaigns five, and the crossing campaigns five on A100-SXM4 and four on RTX A6000.

\textbf{Timed reads.} In the timed selection, the buffer holds as many entries as the scored selection keeps, gathered from the filled cache, and which positions it holds does not change the bytes a step reads. On the timed decode steps the transferred thresholds keep 51\% of the projection bytes at the nominal 50\%, and the fraction varies by less than one point from 2K to 64K.

\textbf{Other timing setups.} The attention-path comparison times dense decoding and the window, whose reads do not depend on cache contents, in one process over allocated caches. Decode-only timing at short context uses a fixed six-token prompt and generates 100 to 200 tokens after warm-up generations, and its Llama rows average five paired dense and sparse launches. End-to-end timing prefills the stated prompt, generates 50 tokens, and reports throughput over the whole interval. The dense batch sweep adds an H100 NVL 94~GB to both GPU models.

\textbf{Quality protocols.} For task scores and chunked perplexity, the model runs as plain completion without the chat template, so only paired differences between dense and sparse variants are meaningful. MMLU~\citep{mmlu} uses 5-shot prompts through the evaluation harness~\citep{lmevalharness2024} on all 14{,}042 questions and sparsifies only the final position of each scoring call. Chunked WikiText-2 perplexity uses 128 chunks of 2{,}048 tokens from the raw test split, processes the first 1{,}024 positions of each chunk densely and the final 1{,}024 sparsely, and pools all next-token losses. Decode-position perplexity instead takes eight evenly spaced windows of the stated length from the WikiText-2 test stream, prefills all but the last 512 tokens densely, and scores those 512 tokens one at a time under the branch, inside the timing harness with the RoPE frequency scaling of Llama-3.1 enabled.

\textbf{Retrieval tests.} The passkey test~\citep{landmark2023} embeds a random six-digit code in filler text at five relative positions, eight trials each, applies the chat template, prefills densely, and decodes eight tokens. A placement counts as inside the window when any needle token lies among the retained sink or recent positions. The multi-key test hides four access codes, each tied to a different named archive, in WikiText-2 test text, asks for the code of one archive, and places the queried code at the same five depths, eight trials each, with the three other codes at random depths. It follows the passkey protocol otherwise, and an answer that gives another archive's code counts as a distractor answer.

\textbf{Scored selection.} The selection re-runs the last 64 prompt tokens after prefill, sums their attention over the cache for each KV head with the query heads of a group pooled and a width-7 average over positions, forces the sinks and the observation window in, and keeps the top positions of each KV head up to the budget of the window, less the tokens that the protocol generates, so that together with the generated tokens it never attends more entries than the window. The scoring pass takes about 0.3~s for a 32K-token prompt on RTX A6000 and 0.23~s on A100-SXM4, and the gather about 10~ms.

\textbf{Threshold transfer.} Thresholds recalibrated on Llama-3.1-8B-Instruct itself change little. At 70\%, 60\%, and 50\% projection keep they lower the perplexity increase from 0.18, 0.40, and 1.00 to 0.16, 0.39, and 0.92, in five, six, and six of the eight windows, with overlapping 95\% intervals. On 32K tokens of teacher-forced text they keep 51.6\% of the projection bytes at the nominal 50\%, against 52.2\% for the transferred thresholds, and every layer stays between 51\% and 53\%. The transferred thresholds are therefore a slightly conservative stand-in for thresholds fitted to the evaluated model.

\textbf{Quality-matched crossover.} Decode-position perplexity uses the protocol above at 32K, with projection keep ratios of 70\%, 60\%, 50\%, and 40\% and selection keep ratios of 70\%, 50\%, 30\%, 20\%, and 10\%. For a budget $b$, each branch removes the largest fraction whose perplexity increase, interpolated linearly between tested keep ratios, stays within $b$. A budget above the whole tested curve takes its most aggressive tested keep ratio, and no curve is extended toward dense decoding. The quality-matched crossover then equates the byte savings of the two admissible fractions with Equation~\ref{eq:crossover}. From a budget of 0.2 upward the selection admits its 10\% keep ratio, which answers every retrieval placement at 32K, so the crossover moves with the admissible projection fraction.

\textbf{Request-mix simulation.} The chat set contains 2{,}000 prompt lengths drawn from a lognormal distribution with log-mean $\log(1024)$ and log-standard-deviation 1.1, clipped to 64 to 32{,}768 tokens, with 256-token answers. The long-document and agent sets each contain 500 uniformly sampled prompt lengths in 32K to 128K and 16K to 96K, with 128-token and 1{,}024-token answers. Mixed requests take the first 1{,}400 chat, 400 long-document, and 200 agent entries.

A request's decode cost integrates the measured latency along its generated sequence in 64-token midpoint segments with log-log interpolation between sweep lengths, clamped at both measured endpoints, and a request that uses the selection is charged its scoring pass in proportion to its prompt length, 0.24~s on A100-SXM4 and 0.32~s on RTX A6000 for 32K tokens. For each request, the gate chooses among dense decoding, the projection branch, the selection, and their composition the configuration whose tabulated cost for the prompt and answer lengths of the request, scoring pass included, is lowest. All policies use the configurations of the context sweep, and the KV window, which fails the retrieval tests, is never chosen.

\section{Projection-Branch Speed and Task Quality}
\label{app:projection}

Table~\ref{tab:app_speed} lists the short-context speed of the projection branch at 50\% keep, decode only and end to end with short prompts.

\begin{table}[H]
\centering
\caption{Projection-branch throughput in tok/s at 50\% keep (FP16, batch size 1, compiled decoding). Llama decode-only rows average five launches with standard deviation at most 2.3\% of the mean, and other rows are single runs.}
\label{tab:app_speed}
\small
\begin{tabularx}{\textwidth}{lYYY}
\toprule
\rowcolor{groupbg}
\textbf{Setting} & \textbf{Dense} & \textbf{Sparse} & \textbf{Speedup} \\
\midrule
\rowcolor{groupyellow}\multicolumn{4}{c}{\textit{Decode only}\strut} \\
Llama-3.1-8B-Instruct, A100 & 98.3 & 134.5 & 1.37$\times$ \\
Llama-3.1-8B-Instruct, RTX A6000 & 44.3 & 75.3 & 1.70$\times$ \\
Mistral-7B-v0.3, RTX A6000 & 48.0 & 82.2 & 1.71$\times$ \\
\rowcolor{groupyellow}\multicolumn{4}{c}{\textit{Prefill and decode, Llama-3.1-8B-Instruct on A100}\strut} \\
Prompt of about 100 words & 70.37 & 87.89 & 1.25$\times$ \\
Prompt of about 500 words & 63.49 & 78.03 & 1.23$\times$ \\
\bottomrule
\end{tabularx}
\end{table}

Table~\ref{tab:app_quality} reports task quality within each task, and Table~\ref{tab:app_sweep} sweeps the keep ratio. Multiple-choice accuracy stays within a point of dense decoding for both variants. Lower keep ratios decode faster on both GPUs as perplexity rises, and the gap between the FF-only and full variants widens with sparsity.

\begin{table}[H]
\centering
\caption{Task quality of the projection branch at 50\% keep on Llama-3.1-8B-Instruct. FF-only sparsifies the feed-forward projections, and Full also sparsifies the attention projections, as the timed branch does.}
\label{tab:app_quality}
\small
\begin{tabularx}{\textwidth}{lYYY}
\toprule
\rowcolor{groupbg}
\textbf{Metric} & \textbf{Dense} & \textbf{FF-only} & \textbf{Full} \\
\midrule
MMLU, 5-shot accuracy (\%) & 68.4 & 67.7 & 67.7 \\
WikiText-2 chunked perplexity & 7.20 & 7.59 & 7.68 \\
\bottomrule
\end{tabularx}
\end{table}

\begin{table}[H]
\centering
\caption{Keep-ratio sweep of the full projection branch. Speed is decode only, from a single launch of each cell, and perplexity uses the chunked protocol for both variants.}
\label{tab:app_sweep}
\small
\begin{tabularx}{\textwidth}{YYYYYYY}
\toprule
\rowcolor{groupbg}
& \multicolumn{2}{c}{\textbf{RTX A6000}} & \multicolumn{2}{c}{\textbf{A100}} & \multicolumn{2}{c}{\textbf{Perplexity}} \\
\rowcolor{groupbg}
\textbf{Keep} & \textbf{tok/s} & \textbf{Speedup} & \textbf{tok/s} & \textbf{Speedup} & \textbf{FF-only} & \textbf{Full} \\
\midrule
100\% & 43.8 & 1.00$\times$ & 98.1 & 1.00$\times$ & 7.20 & 7.20 \\
70\% & 56.7 & 1.29$\times$ & 116.5 & 1.19$\times$ & 7.25 & 7.27 \\
60\% & 63.1 & 1.44$\times$ & 126.2 & 1.29$\times$ & 7.36 & 7.40 \\
50\% & 71.8 & 1.64$\times$ & 135.1 & 1.38$\times$ & 7.59 & 7.68 \\
40\% & 82.7 & 1.89$\times$ & 142.6 & 1.45$\times$ & 8.10 & 8.30 \\
30\% & 97.0 & 2.21$\times$ & 152.6 & 1.56$\times$ & 9.34 & 10.00 \\
\bottomrule
\end{tabularx}
\end{table}

\section{Decode-Step Traffic and Crossover Values}
\label{app:traffic}

With $r_\text{P}{=}0.5$ and $r_\text{KV}{=}0.3$, Equation~\ref{eq:crossover} gives $n^*{=}76{,}069$ tokens for Llama-3.1-8B and Mistral-7B-v0.3 and $n^*{=}67{,}291$ for Qwen3-8B, and the feed-forward projections alone give 61{,}440 and 52{,}663. At 70\% projection keep the Llama-3.1-8B crossover falls to 45{,}641, and at 50\% KV keep it rises to 106{,}496. The layer count cancels, so these values hold for any depth with the same layer dimensions.

Qwen3-8B cannot reach its 51.4K feed-forward crossover inside its 40K window, so Table~\ref{tab:bandwidth} shows how far the cache term approaches the fixed projection term. At 512 tokens the feed-forward weights account for 71.5\% of reads and the cache for 0.5\%. By 32K the cache share is 24.2\%, and the KV saving reaches 62\% of the feed-forward saving. Percent columns follow Equation~\ref{eq:bandwidth} for the deployed BF16 model, with the output head, one embedding row, and all normalization weights in Other, which totals 1.245~GB.

\begin{table}[H]
\centering
\caption{Qwen3-8B decode-step accounting in BF16. The last two columns give the megabytes that the FF-only branch at 50\% keep and the KV branch at 30\% keep remove from each step.}
\label{tab:bandwidth}
\small
\begin{tabularx}{\textwidth}{YYYYYYY}
\toprule
\rowcolor{groupbg}
& \multicolumn{4}{c}{\textbf{Share of step reads (\%)}} & \multicolumn{2}{c}{\textbf{Saved (MB)}} \\
\rowcolor{groupbg}
\textbf{Context} & \textbf{MLP} & \textbf{Attn} & \textbf{KV} & \textbf{Other} & \textbf{MLP} & \textbf{KV} \\
\midrule
512 & 71.5 & 19.9 & 0.5 & 8.2 & 5436 & 53 \\
1K & 71.1 & 19.8 & 1.0 & 8.1 & 5436 & 106 \\
2K & 70.4 & 19.6 & 2.0 & 8.1 & 5436 & 211 \\
4K & 69.1 & 19.2 & 3.8 & 7.9 & 5436 & 423 \\
8K & 66.5 & 18.5 & 7.4 & 7.6 & 5436 & 846 \\
16K & 61.9 & 17.2 & 13.8 & 7.1 & 5436 & 1691 \\
32K & 54.4 & 15.1 & 24.2 & 6.2 & 5436 & 3382 \\
\bottomrule
\end{tabularx}
\end{table}

\section{Harness Costs and Attention Paths}
\label{app:harness}

\textbf{Kernel costs.} For every cell of the two context sweeps we subtract the ideal byte time $t_D B_b/B_D$ from the branch time $t_b$, where $t_D$ is the dense time of the same block and context and the byte counts include the head and norms, and average the difference over blocks. Table~\ref{tab:residuals} lists these kernel costs. On A100-SXM4 the projection branch spends 1.9 to 2.3~ms beyond its byte time at every length, while the selection stays within 0.8~ms of its byte time. The window spends 1.0~ms at 2K and 2.5~ms from 96K, because its second attention call and its offset read into the full cache cost more as context grows. On RTX A6000, whose step takes about twice as long, the projection branch spends 1.0 to 1.5~ms and the selection at most 0.3~ms. The costs also absorb small deviations from nominal read counts, such as the 51\% of projection bytes that the thresholds keep at a nominal 50\%.

\textbf{Cost-aware crossings.} A kernel-cost difference $\Delta c$ between the branches moves their latency crossing away from $n^*$ by $\Delta c/s$, where $s$ is the growth of the KV-branch byte saving as the context gains 1K tokens, the KV bytes that the branch skips within those tokens divided by the effective bandwidth of dense decoding. At 30\% KV keep this is about 0.06~ms on A100-SXM4, whose dense steps move 1.55~TB/s, and 0.13~ms on RTX A6000, whose dense steps move 0.72~TB/s. To predict a crossing we model each branch as $t_b(n) = t_D(n) B_b(n)/B_D(n) + c_b(n)$, interpolate $t_D$ log-log and each kernel cost $c_b$ linearly over the sweep contexts, and solve $t_\text{P}(n) = t_\text{KV}(n)$.

A keep ratio that the sweep lacks takes its cost at 32K from the keep-ratio campaign and the context dependence of the swept cell of the same branch, and each GPU takes these costs from its own sweep and keep-ratio campaign. Kernel costs of A100-SXM4 combined with the dense step time of the RTX A6000 sweep place the RTX A6000 crossings at 38.5K and 65.2K. For Llama-3.2-3B a separate sweep times dense decoding and both branches at 4K, 8K, 16K, 32K, and 48K in three blocks, and its kernel costs follow the same pattern, 1.2 to 1.5~ms for the projection branch and at most 0.44~ms for the selection. No cell of a crossing campaign enters these predictions.

\begin{table}[H]
\centering
\caption{Kernel cost beyond the byte time in ms, the measured step time minus the dense step time scaled by the byte ratio, averaged over the three blocks of each context sweep of Llama-3.1-8B after a real-text prefill.}
\label{tab:residuals}
\small
\begin{tabularx}{\textwidth}{cYYYYYYYY}
\toprule
\rowcolor{groupbg}
& \multicolumn{4}{c}{\textbf{A100-SXM4}} & \multicolumn{4}{c}{\textbf{RTX A6000}} \\
\rowcolor{groupbg}
\textbf{Context} & \textbf{Proj.} & \textbf{Sel.} & \textbf{Window} & \textbf{Both} & \textbf{Proj.} & \textbf{Sel.} & \textbf{Window} & \textbf{Both} \\
\midrule
2K & 2.05 & $-$0.13 & 0.98 & 2.17 & 1.29 & 0.02 & 0.66 & 1.82 \\
4K & 1.95 & $-$0.16 & 1.02 & 2.17 & 1.01 & $-$0.08 & 0.70 & 2.06 \\
8K & 2.13 & $-$0.03 & 1.53 & 2.29 & 1.53 & $-$0.03 & 0.55 & 1.89 \\
16K & 2.25 & 0.14 & 2.01 & 2.36 & 1.18 & 0.17 & 0.66 & 1.92 \\
32K & 2.11 & 0.34 & 2.08 & 2.35 & 1.10 & 0.20 & 0.81 & 2.11 \\
64K & 1.97 & 0.74 & 2.38 & 2.50 & 1.04 & 0.26 & 0.89 & 1.72 \\
96K & 1.95 & 0.43 & 2.52 & 2.41 & -- & -- & -- & -- \\
128K & 1.88 & 0.47 & 2.52 & 2.42 & -- & -- & -- & -- \\
\bottomrule
\end{tabularx}
\end{table}

\textbf{Attention paths.} Table~\ref{tab:kernels} lists the throughput behind the attention-path comparison. The masked kernel is the one \texttt{gpt-fast} ships, the fused kernel is an unmasked call of the same attention primitive, and the split-K kernel is \texttt{flash\_attn\_with\_kvcache}. The retention policy is fixed while its implementation follows the path, so at 128K the window itself runs at 21.6, 21.6, and 63.5~tok/s on the three paths. Peak-bandwidth shares multiply the 32.2~GB that Equation~\ref{eq:bandwidth} assigns to a 128K step by the measured dense throughput and divide by the A100 peak of 2{,}039~GB/s.

\begin{table}[H]
\centering
\caption{Dense throughput in tok/s and KV-window speedup over dense decoding under three compiled attention paths on A100-SXM4, with the byte bound of the window.}
\label{tab:kernels}
\small
\begin{tabularx}{\textwidth}{YYYYYYYY}
\toprule
\rowcolor{groupbg}
& \multicolumn{3}{c}{\textbf{Dense throughput (tok/s)}} & \multicolumn{3}{c}{\textbf{KV-window speedup}} & \\
\rowcolor{groupbg}
\textbf{Context} & \textbf{Masked} & \textbf{Fused} & \textbf{Split-K} & \textbf{Masked} & \textbf{Fused} & \textbf{Split-K} & \textbf{Bound} \\
\midrule
2K & 67.7 & 83.4 & 94.0 & 1.37 & 1.11 & 0.90 & 1.01 \\
4K & 52.3 & 72.3 & 92.8 & 1.69 & 1.22 & 0.92 & 1.02 \\
8K & 36.1 & 60.0 & 89.7 & 2.25 & 1.35 & 0.92 & 1.05 \\
16K & 22.0 & 45.0 & 84.8 & 3.21 & 1.57 & 0.95 & 1.10 \\
32K & 12.1 & 28.1 & 76.3 & 4.66 & 2.00 & 1.02 & 1.18 \\
64K & 6.1 & 14.7 & 64.2 & 6.20 & 2.59 & 1.11 & 1.34 \\
96K & 4.1 & 10.0 & 55.8 & 6.72 & 2.77 & 1.21 & 1.48 \\
128K & 3.1 & 7.7 & 49.2 & 6.92 & 2.82 & 1.29 & 1.60 \\
\bottomrule
\end{tabularx}
\end{table}

\section{Crossing Campaigns}
\label{app:replication}

Each crossing campaign runs after a real-text prefill and times the projection branch and the KV selection of one keep-ratio pair at every context of its grid, in five blocks on A100-SXM4 80~GB GPUs and in four blocks on RTX A6000 48~GB GPUs. Each block holds all cells of a pair on one GPU in fresh processes with full-graph compilation, five warm-up steps, and five repeats of 50 steps, alternates the direction of the context sweep between blocks, and rotates the order of the two cells. Each grid brackets the cost-aware prediction of its pair. For Llama-3.1-8B on A100-SXM4 the contexts come from two launches with the same protocol, and every difference stays paired within its own launch and block. A crossing interpolates the mean difference linearly between the last context where the projection branch leads and the first where the selection leads, and its 95\% interval enumerates all resamples of the blocks, 3{,}125 for five blocks and 256 for four. Tables~\ref{tab:crossing_replication} and~\ref{tab:crossing_replication_a6000} list every context and each crossing with its interval.

\begin{table}[H]
\centering
\caption{Fresh-process step latency in ms on A100-SXM4, averaged over five blocks, with paired differences. Positive differences favor the KV selection, intervals enumerate all block resamples, and the last column counts the blocks in which the selection is faster.}
\label{tab:crossing_replication}
\small
\begin{tabularx}{\textwidth}{YYYYcY}
\toprule
\rowcolor{groupbg}
\textbf{Context} & \textbf{Projection} & \textbf{Selection} & \textbf{Difference} & \textbf{95\% interval} & \textbf{Faster} \\
\midrule
\rowcolor{groupyellow}\multicolumn{6}{c}{\textit{Llama-3.1-8B, keep 0.7\,/\,0.3, byte crossover 44.6K, crossing 25.1K [23.6K, 28.7K]}\strut} \\
20K & 10.65 & 10.98 & $-$0.328 & [$-$0.409, $-$0.255] & 0/5 \\
24K & 11.13 & 11.14 & $-$0.009 & [$-$0.062, 0.032] & 3/5 \\
28K & 11.53 & 11.50 & 0.025 & [$-$0.063, 0.112] & 3/5 \\
32K & 11.87 & 11.66 & 0.212 & [0.126, 0.337] & 5/5 \\
36K & 12.24 & 11.63 & 0.608 & [0.508, 0.687] & 5/5 \\
40K & 12.64 & 12.00 & 0.638 & [0.602, 0.685] & 5/5 \\
44K & 12.88 & 12.10 & 0.781 & [0.721, 0.853] & 5/5 \\
48K & 13.22 & 12.09 & 1.125 & [1.051, 1.199] & 5/5 \\
52K & 13.60 & 12.18 & 1.418 & [1.285, 1.551] & 5/5 \\
\rowcolor{groupyellow}\multicolumn{6}{c}{\textit{Llama-3.1-8B, keep 0.5\,/\,0.3, byte crossover 74.3K, crossing 50.4K [47.4K, 52.5K]}\strut} \\
36K & 10.75 & 11.42 & $-$0.666 & [$-$0.780, $-$0.530] & 0/5 \\
40K & 11.16 & 11.52 & $-$0.356 & [$-$0.415, $-$0.296] & 0/5 \\
44K & 11.48 & 11.73 & $-$0.245 & [$-$0.371, $-$0.119] & 0/5 \\
48K & 11.72 & 11.74 & $-$0.021 & [$-$0.147, 0.067] & 3/5 \\
52K & 12.03 & 12.02 & 0.014 & [$-$0.109, 0.205] & 1/5 \\
56K & 13.03 & 12.36 & 0.670 & [0.556, 0.788] & 5/5 \\
60K & 12.96 & 12.50 & 0.453 & [0.274, 0.632] & 5/5 \\
64K & 13.33 & 12.45 & 0.882 & [0.775, 0.976] & 5/5 \\
68K & 13.71 & 12.80 & 0.913 & [0.862, 0.963] & 5/5 \\
72K & 14.08 & 12.70 & 1.384 & [1.264, 1.504] & 5/5 \\
76K & 14.43 & 12.92 & 1.511 & [1.456, 1.584] & 5/5 \\
80K & 14.76 & 12.92 & 1.839 & [1.735, 1.933] & 5/5 \\
\rowcolor{groupyellow}\multicolumn{6}{c}{\textit{Llama-3.1-8B, keep 0.5\,/\,0.5, byte crossover 104.0K, crossing 68.3K [66.9K, 70.4K]}\strut} \\
56K & 12.83 & 13.00 & $-$0.169 & [$-$0.327, $-$0.037] & 1/5 \\
64K & 13.03 & 13.23 & $-$0.199 & [$-$0.308, $-$0.119] & 0/5 \\
72K & 13.82 & 13.65 & 0.170 & [0.043, 0.297] & 4/5 \\
80K & 14.47 & 13.99 & 0.476 & [0.295, 0.659] & 5/5 \\
88K & 15.43 & 14.48 & 0.954 & [0.869, 1.035] & 5/5 \\
96K & 16.14 & 14.98 & 1.161 & [1.071, 1.251] & 5/5 \\
104K & 16.81 & 15.41 & 1.398 & [1.266, 1.530] & 5/5 \\
112K & 17.52 & 15.64 & 1.880 & [1.729, 1.993] & 5/5 \\
120K & 18.12 & 16.16 & 1.966 & [1.709, 2.210] & 5/5 \\
128K & 18.89 & 16.23 & 2.659 & [2.467, 2.816] & 5/5 \\
\rowcolor{groupyellow}\multicolumn{6}{c}{\textit{Llama-3.2-3B, keep 0.5\,/\,0.3, byte crossover 34.3K, crossing 16.5K [15.6K, 17.2K]}\strut} \\
4K & 4.72 & 5.22 & $-$0.498 & [$-$0.586, $-$0.431] & 0/5 \\
8K & 5.03 & 5.43 & $-$0.402 & [$-$0.441, $-$0.349] & 0/5 \\
12K & 5.36 & 5.51 & $-$0.145 & [$-$0.180, $-$0.105] & 0/5 \\
16K & 5.63 & 5.65 & $-$0.019 & [$-$0.057, 0.014] & 1/5 \\
20K & 5.89 & 5.75 & 0.137 & [0.109, 0.161] & 5/5 \\
24K & 6.24 & 5.83 & 0.407 & [0.371, 0.440] & 5/5 \\
28K & 6.52 & 5.96 & 0.562 & [0.480, 0.618] & 5/5 \\
32K & 6.81 & 6.05 & 0.756 & [0.710, 0.797] & 5/5 \\
\bottomrule
\end{tabularx}
\end{table}

\begin{table}[H]
\centering
\caption{Fresh-process step latency in ms on RTX A6000, averaged over four blocks, with paired differences as in Table~\ref{tab:crossing_replication}.}
\label{tab:crossing_replication_a6000}
\small
\begin{tabularx}{\textwidth}{YYYYcY}
\toprule
\rowcolor{groupbg}
\textbf{Context} & \textbf{Projection} & \textbf{Selection} & \textbf{Difference} & \textbf{95\% interval} & \textbf{Faster} \\
\midrule
\rowcolor{groupyellow}\multicolumn{6}{c}{\textit{Llama-3.1-8B, keep 0.7\,/\,0.3, byte crossover 44.6K, crossing 39.3K [38.9K, 40.3K]}\strut} \\
28K & 22.56 & 24.09 & $-$1.529 & [$-$1.587, $-$1.451] & 0/4 \\
32K & 23.22 & 24.35 & $-$1.131 & [$-$1.232, $-$1.029] & 0/4 \\
36K & 23.94 & 24.50 & $-$0.562 & [$-$0.718, $-$0.402] & 0/4 \\
40K & 24.92 & 24.80 & 0.115 & [$-$0.048, 0.277] & 3/4 \\
44K & 25.60 & 25.04 & 0.562 & [0.524, 0.600] & 4/4 \\
48K & 26.30 & 25.21 & 1.095 & [1.014, 1.215] & 4/4 \\
52K & 27.11 & 25.54 & 1.568 & [1.445, 1.680] & 4/4 \\
\rowcolor{groupyellow}\multicolumn{6}{c}{\textit{Llama-3.1-8B, keep 0.5\,/\,0.3, byte crossover 74.3K, crossing 68.7K [68.5K, 68.9K]}\strut} \\
56K & 24.21 & 25.63 & $-$1.416 & [$-$1.488, $-$1.344] & 0/4 \\
60K & 24.64 & 25.83 & $-$1.191 & [$-$1.239, $-$1.154] & 0/4 \\
64K & 25.37 & 26.05 & $-$0.681 & [$-$0.742, $-$0.625] & 0/4 \\
68K & 26.12 & 26.31 & $-$0.186 & [$-$0.227, $-$0.145] & 0/4 \\
72K & 27.45 & 26.52 & 0.932 & [0.752, 1.046] & 4/4 \\
76K & 27.98 & 26.75 & 1.222 & [1.155, 1.287] & 4/4 \\
80K & 28.64 & 26.93 & 1.707 & [1.616, 1.808] & 4/4 \\
84K & 29.10 & 27.20 & 1.901 & [1.796, 2.030] & 4/4 \\
\bottomrule
\end{tabularx}
\end{table}

\section{Batching}
\label{app:batch}

\textbf{Channel union.} We decode 64 WikiText-2 sequences of Llama-3.1-8B-Instruct at 4K context for 16 steps with the transferred thresholds, split them into groups of $B$ sequences, and record at every step the fraction of the projection weights that at least one sequence of a group activates, which a batched sparse projection must read. Table~\ref{tab:union} averages this union over steps and groups and gives the batched crossover $n^*_B$ of Section~\ref{sec:rule} at 30\% KV keep. At $B{=}1$ the union is the fraction that a single sequence reads, 0.52 on these 4K-context steps rather than the nominal 0.5, so $n^*_1$ lies slightly below the byte crossover of Table~\ref{tab:crossover}.

\begin{table}[H]
\centering
\caption{Fraction of projection weights read by a batched sparse projection, the union of the channels its sequences activate, and the batched crossover at 30\% KV keep for Llama-3.1-8B-Instruct.}
\label{tab:union}
\small
\begin{tabularx}{\textwidth}{lYYYYYY}
\toprule
\rowcolor{groupbg}
\textbf{Batch size $B$} & \textbf{1} & \textbf{2} & \textbf{4} & \textbf{8} & \textbf{16} & \textbf{64} \\
\midrule
Union at 50\% keep & 0.52 & 0.75 & 0.92 & 0.985 & 0.997 & 0.999 \\
Union at 70\% keep & 0.71 & 0.91 & 0.986 & 0.998 & 0.999 & 0.999 \\
$n^*_B$ at 50\% keep & 71.0K & 18.5K & 3.0K & 0.27K & 0.02K & $<$0.01K \\
$n^*_1/B$ at 50\% keep & 71.0K & 35.5K & 17.8K & 8.9K & 4.4K & 1.1K \\
\bottomrule
\end{tabularx}
\end{table}

\textbf{Dense batch sweep.} The batch sweep in Table~\ref{tab:batch} uses the Hugging Face generation loop with eager attention in FP16 rather than the compiled decoding loop, so only the scaling across $B$ is comparable with the main results. Throughput grows 22 to 25 times from one to 64 sequences on all three GPUs while token latency grows less than three times, which is the weight-reuse regime that the batched crossover describes.

\begin{table}[H]
\centering
\caption{Dense decode throughput, token latency, and scaling relative to $B{=}1$ for Llama-3.1-8B-Instruct in the Hugging Face eager loop, with a 128-token prompt and 64 generated tokens.}
\label{tab:batch}
\small
\begin{tabularx}{\textwidth}{cYYYYYYYYY}
\toprule
\rowcolor{groupbg}
& \multicolumn{3}{c}{\textbf{RTX A6000 (48~GB)}} & \multicolumn{3}{c}{\textbf{A100 (80~GB)}} & \multicolumn{3}{c}{\textbf{H100 NVL (94~GB)}} \\
\rowcolor{groupbg}
$B$ & \textbf{tok/s} & \textbf{ms} & \textbf{Scale} & \textbf{tok/s} & \textbf{ms} & \textbf{Scale} & \textbf{tok/s} & \textbf{ms} & \textbf{Scale} \\
\midrule
1 & 35.9 & 27.8 & 1.0$\times$ & 58.8 & 17.0 & 1.0$\times$ & 39.9 & 25.0 & 1.0$\times$ \\
4 & 133.5 & 30.0 & 3.7$\times$ & 208.7 & 19.2 & 3.6$\times$ & 147.3 & 27.2 & 3.7$\times$ \\
8 & 242.5 & 33.0 & 6.8$\times$ & 372.8 & 21.5 & 6.3$\times$ & 263.0 & 30.4 & 6.6$\times$ \\
16 & 409.6 & 39.1 & 11.4$\times$ & 621.6 & 25.7 & 10.6$\times$ & 455.6 & 35.1 & 11.4$\times$ \\
32 & 632.8 & 50.6 & 17.6$\times$ & 958.1 & 33.4 & 16.3$\times$ & 722.9 & 44.3 & 18.1$\times$ \\
64 & 863.6 & 74.1 & 24.1$\times$ & 1285.3 & 49.8 & 21.9$\times$ & 1001.4 & 63.9 & 25.1$\times$ \\
\bottomrule
\end{tabularx}
\end{table}

\section{Long-Context and Two-Turn Quality}
\label{app:longquality}

\textbf{Protocols.} The retrieval tests at 64K and at 127K, which stands for 130{,}048 tokens, follow the 32K protocol of Appendix~\ref{app:details} with the same five depths and eight trials at each depth, and the dense prefill runs through FlashAttention in chunks. The two-turn test starts from a multi-key prompt of 32K or 64K tokens whose first question asks for the code of one archive placed at a random depth, and at 64K its prefill runs through FlashAttention in chunks of 512 tokens. The correct code follows as the first answer, and a second user turn asks for the code of another archive, which sits at the five depths of the single-turn test, eight trials each. The kept selection scores the cache once after the first question and appends every token of the first answer and the second turn to its buffer, whereas the rescored selection scores the whole cache again with the last 64 tokens of the second question. Decode-position perplexity at 64K and 127K follows the 32K protocol with eight windows of each length.

\textbf{Results.} At 127K, dense decoding and the selections at 30\% and 20\% keep miss the same four multi-key placements, two at depth 0.05 and two at depth 0.75, and return the same distractor code in each of them, so the selection reproduces dense decoding even where dense decoding errs. In the two-turn test, dense decoding answers all 40 second questions at both 32K and 64K, and so do the selections at 30\% and 20\% keep and the composition of 50\% projection keep with 20\% selection, whether the selection is kept from the first turn or rescored for the second. Decode-position perplexity of dense decoding is 6.52 at 64K and 6.90 at 127K, and the selection at 30\% and 20\% keep stays within 0.007 of it at both lengths.

\section{Composition Grid and Independent Timing}
\label{app:composition}

The grid crosses projection keep ratios $\{0.7,0.6,0.5\}$ with selection keep ratios $\{0.5,0.3,0.2\}$ and adds the six single-branch settings and dense decoding. All 16 quality evaluations use the same Llama-3.1-8B-Instruct checkpoint, transferred TEAL thresholds, tokenizer, and eight deterministic, non-overlapping 32K windows from the WikiText-2 test stream, whose starts span token positions 0 to 256{,}305. Dense prefill processes 32{,}256 tokens, and each branch then scores the next 512 autoregressively with teacher forcing, giving 4{,}096 scored targets. The KV budgets are 16{,}392, 9{,}835, and 6{,}556 tokens including four sinks, dense aggregate perplexity is 6.0907, and Table~\ref{tab:composition_grid} reports every outcome together with its passkey accuracy at 32K.

The timing screen runs each setting once in the protocol of Appendix~\ref{app:details} on RTX A6000. Before timing, we fixed budgets of 0.1, 0.2, 0.5, and 1.0 in observed aggregate perplexity increase. At each budget we select the fastest eligible setting and the fastest eligible single branch from the screen, which yields seven unique sparse settings, and we re-time them with dense decoding in five blocks on RTX A6000 and in five blocks on A100-SXM4. All seven settings answer every placement of both retrieval tests at 32K. Table~\ref{tab:composition_strict} gives the mean paired speedups. On RTX A6000 the ratios of the selected compositions to their single-branch comparators are 1.250 [1.243, 1.259], 1.264 [1.259, 1.269], and 1.252 [1.231, 1.267] at budgets of 0.2, 0.5, and 1.0. On A100, where the selection alone outruns the 70\% and 60\% projection settings, the fastest single branch within the budgets of 0.2 and 0.5 is the selection, and the three ratios are 1.140 [1.121, 1.159], 1.227 [1.217, 1.235], and 1.237 [1.230, 1.245].

Quality intervals resample the eight paired window losses 20{,}000 times and recompute aggregate perplexity differences. Timing intervals enumerate all $5^5$ resamples of paired launch blocks. Both are 95\% percentile intervals, and configuration selection uses the point estimates.

\begin{table}[H]
\centering
\caption{Complete 32K composition grid with the attention-scored selection as the KV branch. Speedup is the screening ratio to dense decoding on RTX A6000, and a keep ratio of 1 disables the corresponding branch.}
\label{tab:composition_grid}
\small
\begin{tabularx}{\textwidth}{YYYYcYY}
\toprule
\rowcolor{groupbg}
$r_\text{P}$ & $r_\text{KV}$ & \textbf{Perplexity} & \textbf{Increase} & \textbf{95\% interval} & \textbf{Speedup} & \textbf{Passkey} \\
\midrule
1.0 & 1.0 & 6.091 & 0.000 & [0.000, 0.000] & 1.000 & 40/40 \\
0.7 & 1.0 & 6.271 & 0.180 & [0.136, 0.229] & 1.197 & 40/40 \\
0.6 & 1.0 & 6.492 & 0.402 & [0.308, 0.491] & 1.298 & 40/40 \\
0.5 & 1.0 & 7.086 & 0.995 & [0.799, 1.209] & 1.465 & 40/40 \\
1.0 & 0.5 & 6.084 & $-$0.007 & [$-$0.023, 0.006] & 1.108 & 40/40 \\
1.0 & 0.3 & 6.094 & 0.004 & [$-$0.009, 0.018] & 1.151 & 40/40 \\
1.0 & 0.2 & 6.095 & 0.004 & [$-$0.013, 0.019] & 1.182 & 40/40 \\
0.7 & 0.5 & 6.260 & 0.170 & [0.135, 0.205] & 1.383 & 40/40 \\
0.7 & 0.3 & 6.238 & 0.148 & [0.108, 0.181] & 1.440 & 40/40 \\
0.7 & 0.2 & 6.237 & 0.146 & [0.119, 0.171] & 1.490 & 40/40 \\
0.6 & 0.5 & 6.450 & 0.359 & [0.295, 0.418] & 1.529 & 40/40 \\
0.6 & 0.3 & 6.454 & 0.363 & [0.276, 0.461] & 1.576 & 40/40 \\
0.6 & 0.2 & 6.493 & 0.402 & [0.326, 0.466] & 1.663 & 40/40 \\
0.5 & 0.5 & 7.129 & 1.038 & [0.798, 1.305] & 1.637 & 40/40 \\
0.5 & 0.3 & 7.075 & 0.985 & [0.801, 1.182] & 1.736 & 40/40 \\
0.5 & 0.2 & 7.053 & 0.963 & [0.768, 1.196] & 1.795 & 40/40 \\
\bottomrule
\end{tabularx}
\end{table}

\begin{table}[H]
\centering
\caption{Five-block timing of the seven selected sparse settings and dense decoding at 32K, with the byte bound of each setting. Throughput averages across blocks, and speedup averages paired ratios to the dense cell of each block.}
\label{tab:composition_strict}
\small
\setlength{\tabcolsep}{4pt}
\begin{tabular}{cccccc}
\toprule
\rowcolor{groupbg}
& & \multicolumn{2}{c}{\textbf{RTX A6000}} & \multicolumn{2}{c}{\textbf{A100-SXM4}} \\
\rowcolor{groupbg}
$r_\text{P}\,/\,r_\text{KV}$ & \textbf{Bound} & \textbf{tok/s} & \textbf{Speedup [95\% interval]} & \textbf{tok/s} & \textbf{Speedup [95\% interval]} \\
\midrule
1.0\,/\,1.0 & 1.000 & 34.95 & 1.000 [1.000, 1.000] & 74.43 & 1.000 [1.000, 1.000] \\
0.7\,/\,1.0 & 1.277 & 42.65 & 1.220 [1.215, 1.225] & 84.15 & 1.131 [1.118, 1.145] \\
0.6\,/\,1.0 & 1.407 & 46.63 & 1.334 [1.328, 1.341] & 88.81 & 1.193 [1.189, 1.198] \\
0.5\,/\,1.0 & 1.566 & 51.84 & 1.483 [1.478, 1.493] & 94.36 & 1.268 [1.254, 1.282] \\
1.0\,/\,0.2 & 1.217 & 42.11 & 1.205 [1.204, 1.206] & 89.00 & 1.196 [1.189, 1.205] \\
0.7\,/\,0.2 & 1.653 & 53.32 & 1.526 [1.518, 1.537] & 101.43 & 1.364 [1.333, 1.396] \\
0.6\,/\,0.2 & 1.877 & 58.93 & 1.686 [1.681, 1.690] & 109.14 & 1.467 [1.447, 1.487] \\
0.5\,/\,0.2 & 2.172 & 64.90 & 1.857 [1.823, 1.876] & 116.72 & 1.568 [1.559, 1.579] \\
\bottomrule
\end{tabular}
\end{table}

\section{Dispatch on Request Mixes}
\label{app:dispatch}

Table~\ref{tab:dispatch} decomposes the dispatch policies on the mixed workload, priced with the latencies of the sweeps in Figure~\ref{fig:sweep}. The workload combines 70\% chat requests with lognormal prompts of median 1K tokens and 256-token answers, 20\% long-document questions with 32K to 128K prompts and 128-token answers, and 10\% agent loops with 16K to 96K prompts and 1{,}024-token answers. We price each request by integrating the measured latency along its generated sequence, interpolating between the compiled sweep lengths. Always composing and a calibrated gate that may compose at any length each add speed over either single branch, so the gain of dispatch comes from composition.

\end{document}